\documentclass[letterpaper, 10 pt, conference]{ieeeconf}  % Comment this line out if you need a4paper

\IEEEoverridecommandlockouts                              % This command is only needed if 
\usepackage{graphicx}
\usepackage{microtype}

\usepackage{tikz, pgfplots, pgfplotstable}
\usetikzlibrary{calc}
\usetikzlibrary{positioning}
\usepgflibrary{fpu}

\usetikzlibrary{external}
\pgfplotsset{compat=newest}

\usepackage{amsmath}

\usepackage{subcaption}
\DeclareCaptionLabelSeparator{periodspace}{.\quad}
\usepackage{algorithm, algorithmic}
\floatstyle{plaintop}
\restylefloat{algorithm}

\usepackage{placeins}
\usepackage{url}

\usepackage{amsmath}

\newenvironment{customlegend}[1][]{%
	\begingroup
	\csname pgfplots@init@cleared@structures\endcsname
	\pgfplotsset{#1}%
}{%
	\csname pgfplots@createlegend\endcsname
	\endgroup
}%
\def\addlegendimage{\csname pgfplots@addlegendimage\endcsname}

\usepackage{booktabs}

\title{\LARGE \bf Learning Objectness from Sonar Images\\ for Class-Independent Object Detection}

\author{Matias Valdenegro-Toro$^{1}$% <-this % stops a space
    \thanks{$^{1}$Matias Valdenegro-Toro is with the German Research Center for Artificial Intelligence, Robotics Innovation Center. Robert-Hooke-Strasse 1, 28359, Bremen, Germany.
        {\tt\small matias.valdenegro@dfki.de}}%
}

\begin{document}

\maketitle
\thispagestyle{empty}
\pagestyle{empty}

%%%%%%%%%%%%%%%%%%%%%%%%%%%%%%%%%%%%%%%%%%%%%%%%%%%%%%%%%%%%%%%%%%%%%%%%%%%%%%%%
\begin{abstract}
	Detecting novel objects without class information is not trivial, as it is difficult to generalize from a small training set. This is an interesting problem for underwater robotics, as modeling marine objects is inherently more difficult in sonar images, and training data might not be available a priori. Detection proposals algorithms can be used for this purpose but usually requires a large amount of output bounding boxes. In this paper we propose the use of a fully convolutional neural network that regresses an objectness value directly from a Forward-Looking sonar image. By ranking objectness, we can produce high recall (96 \%) with only 100 proposals per image. In comparison, EdgeBoxes requires 5000 proposals to achieve a slightly better recall of 97 \%, while Selective Search requires 2000 proposals to achieve 95 \% recall. We also show that our method outperforms a template matching baseline by a considerable margin, and is able to generalize to completely new objects. We expect that this kind of technique can be used in the field to find lost objects under the sea.
\end{abstract}

%%%%%%%%%%%%%%%%%%%%%%%%%%%%%%%%%%%%%%%%%%%%%%%%%%%%%%%%%%%%%%%%%%%%%%%%%%%%%%%%
\section{Introduction}

Perception is a fundamental problem in Robotics, particularly for underwater environments. Many real-world tasks require a robot to first find an object and then identify it. A more difficult task is to find an object from an abstract description or to find all objects in the scene irrespective if they are from known classes or not.

Finding and detecting objects under water is also a hard problem, mostly due to light absorption by water that degrades the use of optical sensors, and the difficulties associated at interpreting  acoustic sensor outputs. There have been large advances at detecting objects in sonar images \cite{hurtos2013automatic}, but they usually make strong assumptions on object shape or require shadow/highlight segmentation \cite{sawas2010cascade}. Techniques to detect objects if a medium sized training set are available \cite{valdenegro2016submerged}, but many tasks require a robot to find an "novel" object, where training samples are not available a priori. Examples of this are finding submerged airplane wrecks, and detecting marine debris.

This paper deals with the problem of building a class-agnostic object detector for sonar images. In the computer vision literature, this is called detection proposals \cite{hosang2014good}, but in that context they are only used in order to "guide" the object detection process and improve localization results. We believe that detection proposals are also useful in underwater robotics on its own, where the idea is to detect objects, even if their shape or content has not been seen before by the system. In this case, class predictions are not available, but novel objects can still be identified by the detection system as different than background.

This work expands on our previous work \cite{valdenegroToro2016Proposals}, where we introduced a basic version of this system. This work introduces the following contributions:

\begin{itemize}
	\item We propose the use of objectness ranking in order to increase the adaptivity of the system to different environments.
	\item We introduce a new neural network objectness regressor that requires considerably less parameters and is fully convolutional, resulting in a four times improvement in computation time.
	\item We perform a more thorough evaluation and compare with the state of the art, showing that our method outperforms other methods and requires less proposals to achieve high recall.
\end{itemize}

These contributions result in more appropriate technique for underwater robot perception.

\section{Related Work}

The underwater perception literature contains many techniques to detect objects in sonar images. A very popular option is the use of template matching \cite{hurtos2013automatic}, where a set of templates is cross-correlated with the input image, and this produces maximum correlation where the object is located. A threshold is usually set to avoid false positives.

Another option is using classic computer vision methods, like the boosted cascade of weak classifiers \cite{sawas2010cascade}, but this only works well in objects that produce large sonar shadows, as Haar features correlate very well with this feature.

Neural networks have also been used \cite{valdenegro2016submerged}, where a CNN is trained on image patches and used in a sliding window fashion on a test image. This technique works quite well in terms of accuracy but it produces a large amount of false positives. An end-to-end multi-task approach \cite{valdenegro2016end} improves false detections by explicitly modeling the detection process as proposals. 

In general, CNN-based techniques are able to model more complex objects than classic computer vision methods. Template matching in particular is not able to model objects more complex than underwater mines, such as marine debris.

The concept of detection proposals is introduced in the computer vision literature \cite{endres2010category}, where instead of using an expensive sliding window to detect objects, the detection process can be "guided" by a subset of windows that are likely to contain objects. A detection proposals algorithm infers these bounding boxes (also called proposal) from image content. Proposals are also linked to the concept of "objectness" \cite{alexe2012measuring}, where the authors define it as "quantifying how likely it is for an image window to contain an object of any class". A set of predefined cues are combined in order to produce objectness, which can be used to discriminate between object and background windows.

EdgeBoxes \cite{zitnick2014edge} is a proposals technique that uses a structured edge detector, which extracts high quality edges. Edges are then grouped to produce object proposals that can be scored by predefined techniques. This method is very fast but needs a large amount of proposals to produce high recall. Selective Search \cite{uijlings2013selective} takes a different approach, by doing super-pixel segmentation and using a set of strategies to merge super-pixels into detection proposals. It is quite slow but it can achieve very high recall with a medium number of output proposals.

Neural networks have also been used to model detection proposals. The best teachnique is the Region Proposal Network from the Faster R-CNN object detection framework \cite{ren2015faster}. The RPN module regresses bounding box coordinates and outputs a binary decision corresponding to object vs background. The RPN works quite well on color images and improves the state of the art in the PASCAL VOC 2007/2012 datasets, but we have not been able to train such modules for proposals on sonar images, mostly likely due to the small scale datasets that we have.

While there are established techniques for detection proposals in color images, these are not directly transferrable to sonar images. Bounding box regression techniques cannot be trained unless a large dataset ($\sim $ 1M images) is available for pre-training. The typical dataset of sonar images ranges in the thousands, preventing the use of such techniques.
We have developed a simple objectness regressor \cite{valdenegroToro2016Proposals} using neural networks that works well for detection proposals, but it is computationally expensive as features are not shared across neural network evaluations, and simple thresholding of objectness values might not generalize well across environments.

A bigger concern for robot perception is that most detection proposals techniques require a large number of output proposals to achieve high recall. This means that most proposals might not correspond to actual objects in the image, which defeats the purpose of a proposals method over a simple sliding window. In this work we evaluate how high recall can be achieved with a low number of output proposals by ranking objectness instead of using a fixed threshold.

\section{Learning Objectness from Sonar Image Patches}

Objectness is an abstract concept that quantifies the property that an image window contains an object. A window containing an object should be assigned a high objectness score, while a window with only background should receive a low objectness score.

Our method is based on the idea that objectness can be estimated from an image window/patch. Given ground truth objectness values, an objectness regressor can be trained on such data to learn the relationship between image content and an abstract objectness score concept. This corresponds to a data-driven approach.

\textbf{Training Data Generation}. We compute ground truth objectness as follows. We run a $n \times n$ sliding window with a stride of $s$ pixels in each direction, and for each ground truth bounding box in the image, we assign a positive objectness score $o$ to the sliding window that has the highest Intersection-over-Union score (IoU, Eq \ref{iouEquation}). We also assign a positive objectness score to any sliding window with $\text{IoU} \geq 0.5$. This is intended to introduce variety in the range of objectness values.

\begin{equation}
\text{IoU}(A, B) = \frac{\text{area}(A \cap B)}{\text{area}(A \cup B)}
\label{iouEquation}
\end{equation}

The IoU score is commonly used in computer vision to evaluate object detection algorithms \cite{zitnick2014edge} \cite{girshick2015fast}. Typically in order of 5 to 10 windows with positive objectness are generated for each ground truth bounding box. To generate negative objectness windows, we randomly sample $N = 10$ windows that have a maximum IoU with the ground truth bounding boxes of $\epsilon$. All negative objectness windows receive a zero objectness score. Positive and negative windows are cropped and stored as a labeled dataset to train an objectness regressor.

The final ground truth objectness score is obtained as:

\begin{equation}
	\text{objectness}(\text{iou}) = 
	\begin{cases} 
		1.0 		& \text{if iou} \geq 1.0 - \epsilon \\
		\text{iou}  & \text{if } 1.0 - \epsilon < \text{iou} < \epsilon\\
		0.0      	& \text{if } \text{iou} \leq \epsilon
	\end{cases}
	\label{iouObjectness}
\end{equation}

The motivation for using Eq \ref{iouObjectness} is to expand the range of available objectness scores. While the IoU is in the $[0, 1]$ range, obtaining a IoU score close to $1.0$ is very unlikely, as it would imply a near-perfect match between the ground truth and the sliding window. We introduce a tolerance where any IoU bigger than $1.0 - \epsilon$ is considered equivalent as the maximum objectness range. The lower threshold is to remove any window that might not have enough intersection with the ground truth. In our experiments we use $\epsilon = 0.2$. Any IoU value between the lower and upper thresholds is kept directly as the objectness score in order to introduce variability into the ground truth objectness scores.

\textbf{Network Architectures}. We use two CNN models that take a $96 \times 96$ one-channel sonar image patch as input, and output an objectness score in the $[0, 1]$ range. The first model was previously proposed by us \cite{valdenegroToro2016Proposals} and contains approximately 900K parameters. In order to apply this model to full-size sonar image, we used a sliding window. The second model only contains 20K parameters and is fully convolutional, which allows it to take a full-size sonar image and output objectness for each pixel, sharing computation and avoiding computational performance issues. This model also has the advantage of allowing variable-sized images.

We use the following notation. Conv($n$, $s$) a 2D Convolutional module with \textit{n} square filters of size $s$, MP($s$) a Max-Pooling module with subsampling size $s$, FC($n$) a fully connected layer with $n$ output neurons. 

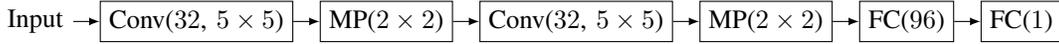
\begin{figure*}[tb]
	\centering
	\begin{tikzpicture}[style={align=center, minimum width=1cm}]
	\node (I) {Input};
	\node[draw, right=1em of I] (conv2d1) {Conv($32$, $5 \times 5$)};
	\node[draw, right=1em of conv2d1] (mp1) {MP($2 \times 2$)};
	\node[draw, right=1em of mp1] (conv2d2) {Conv($32$, $5 \times 5$)};
	\node[draw, right=1em of conv2d2] (mp2) {MP($2 \times 2$)};
	\node[draw, right=1em of mp2] (fc1) {FC($96$)};
	\node[draw, right=1em of fc1] (fc2) {FC($1$)};
	\draw[-latex] (I) -- (conv2d1);
	\draw[-latex] (conv2d1) -- (mp1);
	\draw[-latex] (mp1) -- (conv2d2);
	\draw[-latex] (conv2d2) -- (mp2);
	\draw[-latex] (mp2) -- (fc1);
	\draw[-latex] (fc1) -- (fc2);
	\end{tikzpicture}
	\caption{CNN model based on the LeNet Architecture. All layers use ReLU activation, except the last layer that uses a sigmoid function.}
	\label{fig:classicModel}
\end{figure*}

The first CNN model is based on LeNet \cite{lecun1998gradient}, with two stacks of convolutional and max-pooling layers, and two fully connected layers. The network architecture is shown in Fig. \ref{fig:classicModel}. ReLU is used as activation except at the output layer, where a sigmoid activation is used.

\begin{figure*}[!htb]
	\centering
	\begin{tikzpicture}[style={align=center, minimum width=1cm}]
	\node (I) {Input};
	\node[draw, right=1em of I] (A) {Conv($24, $$3 \times 3$)};
	\node[draw, right=1em of A] (B) {Conv($24, $$1 \times 1$)};
	\node[draw, right=1em of B] (C) {MP($2 \times 2$)};
	\node[draw, right=1em of C] (D) {Conv($24, $$3 \times 3$)};
	\node[draw, right=1em of D] (E) {Conv($24, $$1 \times 1$)};
	\node[draw, right=1em of E] (F) {MP($2 \times 2$)};
	\node[draw, right=1em of F] (out) {FC($1$)};
	\draw[-latex] (I) -- (A);
	\draw[-latex] (A) -- (B);
	\draw[-latex] (B) -- (C);
	\draw[-latex] (C) -- (D);
	\draw[-latex] (D) -- (E);
	\draw[-latex] (E) -- (F);
	\draw[-latex] (F) -- (out);
	\end{tikzpicture}
	\caption{FCN model based on the Tiny module \cite{ValdenegroToro2017RealtimeCN}. All layers use ReLU activation, except the last layer that uses a sigmoid function.}
	\label{fig:fcnModel}
\end{figure*}
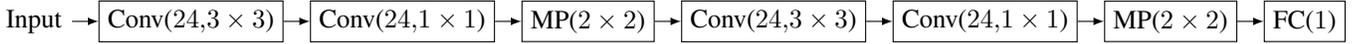

The second model is based on SqueezeNet \cite{iandola2016squeezenet}, from which we have derived the Tiny module \cite{ValdenegroToro2017RealtimeCN} that allows a model with large expressivity and a low number of parameters. The model shown in Fig. \ref{fig:fcnModel} is trained, and for test-time inference, it is modified to construct a fully convolutional version of it. This is done by replacing the last fully connected (FC) layer with a Conv($1$, $24 \times 24$) layer and reshaping the weights \cite{Shelhamer2015FullyCN} to fit convolutional filters. Given an input image, this model produces an output objectness map that is smaller than the input, due to the use of max-pooling. We up-sample the objectness map back to the original input size with bilinear interpolation. The down-scaling factor is defined by the model architecture (Fig. \ref{fig:fcnModel}) and the number of max-pooling layers. Minimizing this scaling factor inherent in the model is what motivates the use of a simplified model, with a single fully connected layer.

\textbf{Training}. Both models are trained in the same way, using a mean squared error loss with the ADAM optimizer \cite{kingma2014adam} and a learning rate $\alpha = 0.01$. Training stops when the loss converges, determined by early stopping on a held-out validation set, which usually happens after 15-20 epochs. No pre-training or fine-tuning is performed, and all weights start from random initialization.

\section{Detection Proposals from Objectness Scores}

We propose two methods to convert objectness scores into dense detection proposals. First, a $96 \times 96$ with stride $s = 4$ sliding window is applied to the input image and objectness scores are computed for each window. Then candidate windows are filtered into detection proposals by a given method:

\begin{itemize}
	\item \textbf{Thresholding}. Any window with objectness bigger than a threshold $T_o$ is output as a proposal. The value of $T_o$ can be tuned in a validation set given a recall target, but in general this parameter determines a trade-off between recall and number of proposals.
	\item \textbf{Ranking}. All candidate windows are sorted by decreasing objectness and the top $k$ ones are output as detection proposals. This method introduces a quality parameter $k$, which is directly related to recall and can be tuned to maximize recall under a given number of proposals.
\end{itemize}

The sliding window approach is only used with the CNN model in order to build an objectness map, while the FCN model implicitly does a sliding window as convolution, and outputs the objectness map directly. After deciding proposal windows, we apply non-maximum suppression with a given threshold $T_s$ in order to reduce duplicate detections. We use $T_s = 0.8$ as a good compromise between number of output proposals and recall. Note that our method can potentially produce proposals at multiple scales, but in this work we only report results with a single scale (given by the $96 \times 96$ window).

\section{Experimental Evaluation}

In this section we evaluate our methods and provide comparisons with the state of the art.

\textbf{Data}. We use a marine debris dataset\footnote{The full dataset is available at \url{https://github.com/mvaldenegro/marine-debris-fls-datasets/releases/}} of 2000 full-sized sonar images obtained from an ARIS Explorer 3000 Forward-Looking sonar. They were captured at the Ocean Systems Lab (Heriot-Watt University) water tank and contain marine debris objects such as cans, bottles, tires, etc. After extracting patches using a sliding window with a stride $s = 4$ from 1300 full-sized sonar images, we obtained 51563 training and 22137 validation samples (70\%/30\% split). The remaining 700 full-size sonar images are used for evaluation and comparison of detection proposal techniques.

\textbf{Metrics}. The typical metric \cite{uijlings2013selective} to evaluate detection proposals is the recall:

\begin{equation}
R = \frac{TP}{TP + FN}
\end{equation}

Where $TP$ is the number of true positives, and $FN$ is the number of false negatives. A proposal is considered a correct if the IoU (Eq \ref{iouEquation}) between proposal and ground truth bounding boxes is greater than some threshold $T_d$. The most common value for this threshold is $T_d = 0.5$.

%\begin{equation}
%\text{bestMatch}(A, GT) = \max_{R \in GT} \text{IoU}(A, R)
%\label{eq:bestMatch}
%\end{equation}

Recall is used because a detection proposal method typically generalizes well and it can generate many bounding boxes that correspond to real objects in the image, but are not labeled as such. Precision is not typically evaluated as unlabeled objects are considered false positives \cite{hosang2015makes} which would skew any evaluation. The area under the ROC curve (AUC) is also not appropriate for the same reasons. Note that this evaluation protocol can be "gamed" due to partially annotated datasets \cite{chavali2016object}.

Another important metric is the number of output bounding boxes (also called proposals), as it is very easy to obtain high recall with a high number of proposals, but too many bounding boxes hurt the applicability of such methods as many boxes do not correspond to real objects. An ideal detection proposals technique would have high recall with a relatively low number of output proposals.

%It should also be mentioned that the use of recall with a partially annotated dataset could be "gamed" and could be overfitting to the objects that are labeled \cite{chavali2016object}. In our evaluation we manually validate that our method also detects other blobs that could be objects but are not labeled, so we are certain that such overfitting is not happening.

\textbf{Baselines}. There are no detection proposal techniques specifically designed for sonar images, which makes defining a baseline not trivial. We compare against our previous work \cite{valdenegroToro2016Proposals}, and we evaluate a number of baselines, namely: EdgeBoxes \cite{zitnick2014edge}, Selective Search \cite{uijlings2013selective}, and Cross-Correlation Template Matching \cite{midelfart2009template}.

EdgeBoxes extracts high-quality edges and groups them into object proposals at multiple scales. A score threshold is required, and the number of output proposals can also be tuned. We evaluate both parameters by selecting a low threshold $0.0001$ at a fixed number of 300 proposals, and using a $0.0$ score threshold and varying the number of output proposals.

Selective Search uses a predefined set of strategies that merge super-pixels into detection proposals. We evaluate both the Quality and Fast configurations, with a variable number of output proposals. For both EdgeBoxes and Selective Search, we used the OpenCV ximproc \footnote{Available at \url{https://github.com/opencv/opencv_contrib/tree/master/modules/ximgproc}} module implementation.

We also built a detection proposals algorithm using a cross-correlation similarity typically used for template matching. We randomly selected a set of $T = 100$ positive patches from the training set and computed the maximum cross-correlation between an input patch and all templates. This works as a pseudo-objectness measure and we use this score for both thresholding and ranking proposals.

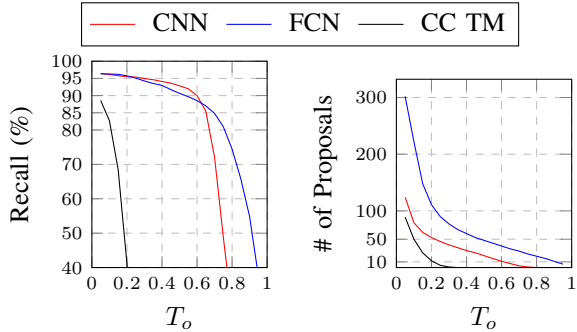
\begin{figure}[!h]
	\centering
	\begin{tikzpicture}
		\begin{customlegend}[legend columns = 3,legend style = {column sep=1ex}, legend cell align = left,
		legend entries={CNN, FCN, CC TM}]
			\addlegendimage{mark=none,red}
			\addlegendimage{mark=none,blue}		
			\addlegendimage{mark=none,black}
		\end{customlegend}
	\end{tikzpicture}
	\begin{subfigure}[b]{0.22 \textwidth}		
		\begin{tikzpicture}
		\begin{axis}[height = 0.18 \textheight, width = \textwidth, xlabel={$T_o$}, ylabel={Recall (\%)}, xmin = 0.0, xmax = 1.0, ymin = 40.0, ymax = 100.0, xtick = {0.0, 0.2, 0.4, 0.6, 0.8, 1.0}, ytick = {40, 50, 60, 70, 80, 85, 90, 95, 100}, ymajorgrids=true, xmajorgrids=true, grid style=dashed, legend pos = north east, legend style={font=\scriptsize}, tick label style={font=\scriptsize}]
		
		\addplot+[mark = none, red] table[x  = threshold, y  = meanRecall, col sep = space] {data/thresholdVsRecallNMS0.80.csv};
		
		\addplot+[mark = none, blue] table[x  = threshold, y  = meanRecall, col sep = space] {data/fcnProposals-thresholdVsRecallNMS0.80.csv};
		
		\addplot+[mark = none, black] table[x  = threshold, y  = meanRecall, col sep = space] {data/tmProposals-TSPC100-thresholdVsRecallNMS0.80.csv};
		
		\end{axis}		
		\end{tikzpicture}
	\end{subfigure}		
	\begin{subfigure}[b]{0.22 \textwidth}		
		\begin{tikzpicture}
		\begin{axis}[height = 0.17 \textheight, width = \textwidth, xlabel={$T_o$}, ylabel={\# of Proposals}, xmin=0.0, xmax = 1.0, ymin = 0.0, xtick = {0.0, 0.2, 0.4, 0.6, 0.8, 1.0}, ytick = {10, 50, 100, 200, 300}, ymajorgrids=true, xmajorgrids=true, grid style=dashed, legend pos = north east, legend style={font=\scriptsize}, tick label style={font=\scriptsize}]
		
		\addplot+[mark = none, red] table[x  = threshold, y  = meanNumProposals, col sep = space] {data/thresholdVsRecallNMS0.80.csv};
		\addplot+[mark = none, blue] table[x  = threshold, y  = meanNumProposals, col sep = space] {data/fcnProposals-thresholdVsRecallNMS0.80.csv};
		\addplot+[mark = none, black] table[x  = threshold, y  = meanNumProposals, col sep = space] {data/tmProposals-TSPC100-thresholdVsRecallNMS0.80.csv};
		
		\end{axis}		
		\end{tikzpicture}
	\end{subfigure}
	\caption{Objectness thresholding results with CNN, FCN and CC TM objectness. CNN performs slightly better than FCN, while CC TM fails to generalize properly.}
	\label{cnnThresholdingPlotResults}
\end{figure}

\textbf{Results}. A comparison between CC template matching and CNN/FCN objectness with thresholding is shown in Fig. \ref{cnnThresholdingPlotResults}. Our results show that CC TM performs quite poorly, while objectness produced by CNN and FCN perform better, with slowly decreasing recall and number of proposals as $T_o$ is increased. FCN performs slightly worse than CNN, indicated by requiring approximately two times the number of proposals to produce the same recall.

\begin{figure}[!h]
	\centering
	\begin{tikzpicture}
	\begin{customlegend}[legend columns = 3,legend style = {column sep=1ex}, legend cell align = left,
	legend entries={CNN, FCN, CC TM}]
	\addlegendimage{mark=none,red}
	\addlegendimage{mark=none,blue}		
	\addlegendimage{mark=none,black}
	\end{customlegend}
	\end{tikzpicture}
		\begin{tikzpicture}
		\begin{axis}[height = 0.17 \textheight, width = 0.45 \textwidth, xlabel={$k$}, ylabel={Recall (\%)}, xmin = 0.0, xmax = 100, ymin = 40.0, ymax = 100.0, ytick = {40, 50, 60, 70, 80, 85, 90, 95, 100}, ymajorgrids=true, xmajorgrids=true, grid style=dashed, legend pos = north east, legend style={font=\scriptsize}, tick label style={font=\scriptsize}]
		
		\addplot+[mark = none, red] table[x  = k, y  = meanRecall, col sep = space] {data/topKVsRecallAtIoU0.50NMS0.80.csv};
		
		\addplot+[mark = none, blue] table[x  = k, y  = meanRecall, col sep = space] {data/fcnProposals-topKVsRecallAtIoU0.50NMS0.80.csv};
		
		\addplot+[mark = none, black] table[x  = k, y  = meanRecall, col sep = space] {data/tmProposals-TSPC100-topKVsRecallAtIoU0.50NMS0.80.csv};
		
		\end{axis}		
		\end{tikzpicture}
		%\caption{Recall vs Threshold}       
	\caption{Objectness ranking results with CNN, FCN, and CC TM objectness. CNN can obtain high recall from 20-40 proposals per image, while FCN requires more to generalize properly. CC TM needs even more proposals to obtain poor recall, which suggests that it is not a good choice for this problem.}
	\label{cnnRankingPlotResults}
\end{figure}
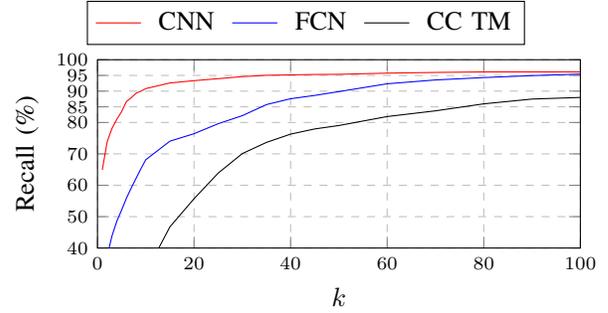

A comparison of objectness ranking is shown in Fig. \ref{cnnRankingPlotResults}. Results show again that CNN produces better objectness, reflected as requiring less proposals, but still FCN objectness can obtain $95$ \% recall with only 100 proposals per image. CC TM objectness saturates at $88$ \% recall if more than 110 proposals are output.

Thresholding and Ranking both have a best recall at $95$ \%, but ranking has the advantage of requiring less output proposals to achieve high recall, 40 for CNN and 80-100 for FCN, while thresholding requires considerable more proposals to achieve the recall target. This indicates that ranking can possibly adapt better to unknown environments, even as the objectness scores change.

Table \ref{comparisonTable} provides a global comparison with the state of the art. For each method, we determined the configuration that produces the best recall, the number of output proposals required to achieve such recall, and computation time as evaluated on a AMD Ryzen 7 1700 processor.

EdgeBoxes is by far the fastest method at $0.1$ seconds to process one frame, and it produces the best recall, but doing so requires 5000 proposals per image. Selective Search Quality also obtains very good recall but with a large number of proposals. CC TM produces the lowest recall we observed on this experiment.

Our proposed techniques obtain very good recall with a low number of proposals per image. CNN-Ranking produces $96$ \% recall with only 80 proposals per image, which is 62 times less than EdgeBoxes with only a $1$ \% absolute loss in recall. Selective Search produces $1$ \% less recall than the best of our methods, but outputting 25 times more proposals.

In terms of computation time, EdgeBoxes is the fastest. FCN objectness is 4 times faster to compute than CNN objectness, due to the fully convolutional network structure, and it only requires a $1$ \% reduction in recall. CC Template Matching is also quite slow, at 10 seconds per image.

\begin{figure*}[ht]
	\centering	
	\begin{subfigure}[b]{0.49 \textwidth}
		\includegraphics[width=0.36\textwidth]{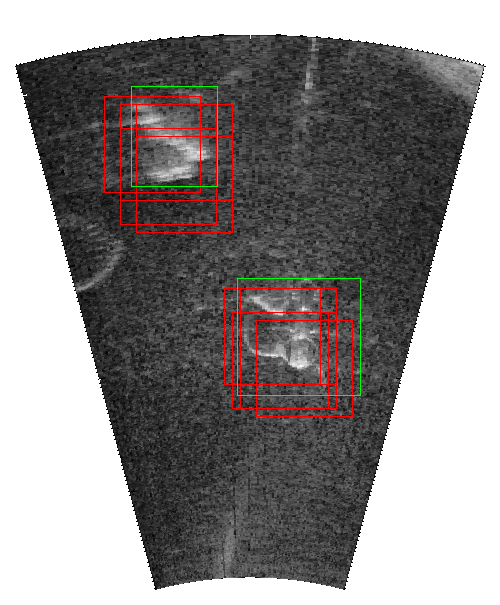}
		\includegraphics[width=0.28\textwidth]{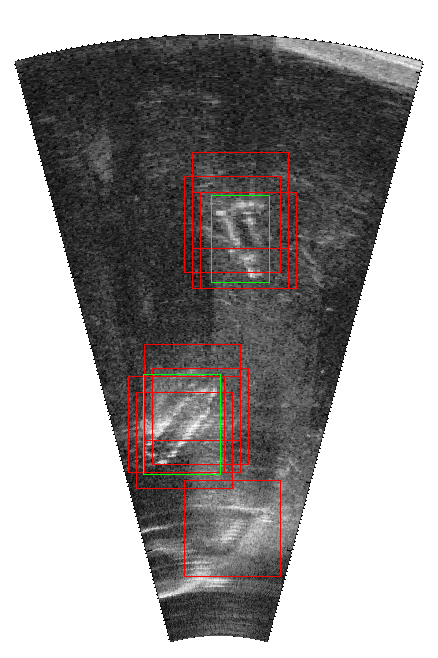}
		\includegraphics[width=0.28\textwidth]{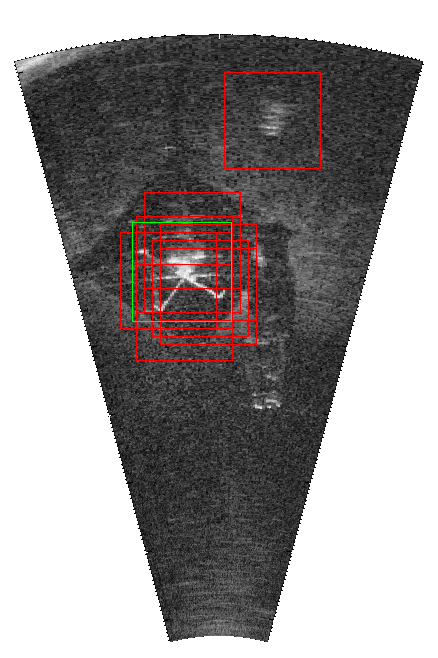}
		\caption{CNN}
	\end{subfigure} \,
	\begin{subfigure}[b]{0.49 \textwidth}
		\includegraphics[width=0.36\textwidth]{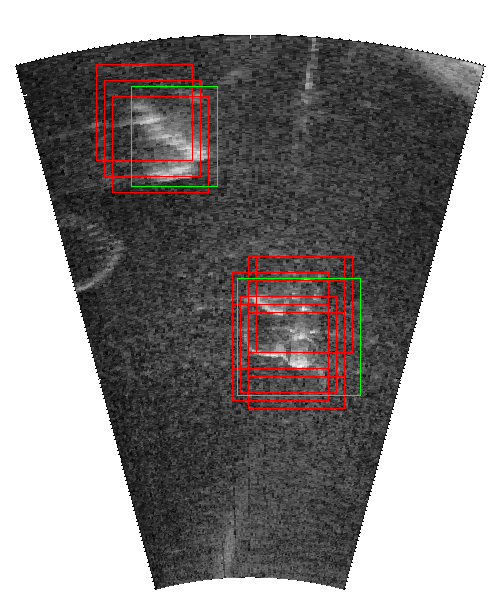}
		\includegraphics[width=0.28\textwidth]{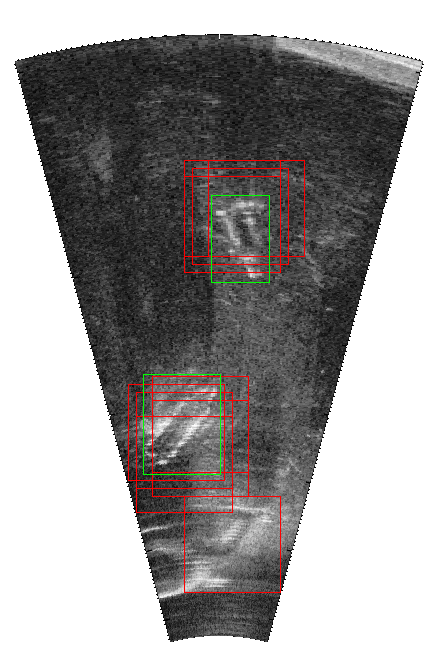}
		\includegraphics[width=0.28\textwidth]{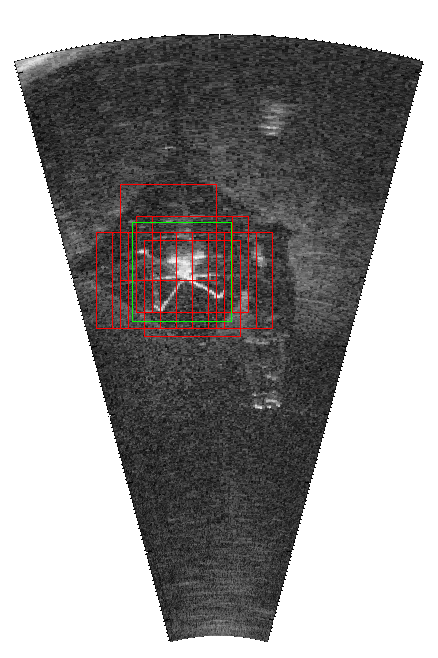}
		\caption{FCN}
	\end{subfigure}
	\caption{Sample detections produced by objectness ranking with CNN and FCN scores. We show the top $K = 10$ scoring detections. Red bounding boxes are detections, while green ones are the ground truth. Note how CNN in some cases detects blob objects that are unlabelled in our dataset.}
	\label{exampleDetections}
\end{figure*}

In Fig. \ref{exampleDetections} we show a small sample of CNN and FCN detections produced by objectness ranking. Our results show that classical sonar object detection techniques are not really appropriate for detections proposals, as they are slow and cannot achieve high recall. This is likely due to inability to model high object variation.

\begin{table}
	\centering
	\begin{tabular}{llll}
		\hline 
		Method 						& Best Recall 		& \# of Proposals & Time (s)\\ 
		\hline
		TM CC Threshold				& $91.83$ \% 	& 150 & $10 \pm 0.5$\\
		TM CC Ranking				& $88.59$ \% 	& 110 & $10 \pm 0.5$\\		
		\hline 
		EdgeBoxes (Thresh)			& $57.01$ \%	& 300	& $0.1$\\ 
		EdgeBoxes (\# Boxes)		& $\mathbf{97.94}$ \%	& 5000	& $0.1$\\
		\hline
		Selective Search Fast		& $84.98$ \%	& 1000	& $1.5 \pm 0.1$\\
		Selective Search Quality	& $\mathbf{95.15}$ \%	& 2000	& $5.4 \pm 0.3$\\
		\hline						 		
		CNN-Threshold				& $\mathbf{96.42}$ \%	& \textbf{125}	& $12.4 \pm 2.0$\\
		FCN-Threshold				& $\mathbf{96.33}$ \%	& 300	& $3.1 \pm 1.0$\\
		CNN-Ranking					& $\mathbf{96.12}$ \%	& \textbf{80}	& $12.4 \pm 2.0$\\
		FCN-Ranking					& $\mathbf{95.43}$ \%	& \textbf{100}	& $3.1 \pm 1.0$\\
		\hline 
	\end{tabular}
	\caption{Comparison of detection proposal techniques on Forward-Looking Sonar Images. Our proposed methods obtain the highest recall with the lowest number of proposals. Only EdgeBoxes has a higher recall with a considerably larger number of output proposals.}
	\label{comparisonTable}
\end{table}

\begin{figure*}
	\centering	
	\begin{subfigure}[c]{0.79 \textwidth}			
	\begin{tikzpicture}
		\begin{axis}[height = 0.22 \textheight, width = \textwidth, xlabel={\# of Proposals}, ylabel={Recall (\%)}, xmin = 1.0, xmax = 10000, ymin = 0.0, ymax = 100.0, ytick = {0, 20, 40, 60, 80, 85, 90, 95, 100}, ymajorgrids=true, xmajorgrids=true, grid style=dashed, legend pos = north east, legend style={font=\scriptsize}, tick label style={font=\scriptsize}, xmode=log]
			%SS Fast
			\addplot+[red, solid, mark = none] table [x = k, y = meanRecall] {data/selectiveSearchFastTopKVsRecallAtIoU0.50.csv};
			
			%SS Quality
			\addplot+[red, dashed, mark = none] table [x = k, y = meanRecall] {data/selectiveSearchQualityTopKVsRecallAtIoU0.50.csv};
		
			%EdgeBoxes
			\addplot+[black, solid, mark = none] table [x = k, y = meanRecall] {data/edgeBoxes-topKVsRecallAtIoU0.50.csv};
		
			%Thresholds
			\addplot+[blue, solid, mark = none] table[x  = meanNumProposals, y = meanRecall, col sep = space] {data/thresholdVsRecallNMS0.80.csv};
			
			\addplot+[blue, dashed, mark = none] table[x  = meanNumProposals, y = meanRecall, col sep = space] {data/fcnProposals-thresholdVsRecallNMS0.80.csv};
			
			\addplot+[blue, densely dotted, thick, mark = none] table[x  = meanNumProposals, y = meanRecall, col sep = space] {data/tmProposals-TSPC100-thresholdVsRecallNMS0.80.csv};
			
			%Ranking
			\addplot+[brown, solid, mark = none] table[x  = k, y = meanRecall, col sep = space] {data/topKVsRecallAtIoU0.50NMS0.80.csv};
			
			\addplot+[brown, dashed, mark = none] table[x  = k, y = meanRecall, col sep = space] {data/fcnProposals-topKVsRecallAtIoU0.50NMS0.80.csv};
			
			\addplot+[brown, densely dotted, thick, mark = none] table[x  = k, y = meanRecall, col sep = space] {data/tmProposals-TSPC100-topKVsRecallAtIoU0.50NMS0.80.csv};
		\end{axis}		
	\end{tikzpicture}
	\end{subfigure}
	\begin{subfigure}[c]{0.20 \textwidth}
		\vspace*{-1.0cm}
		\begin{tikzpicture}
		\begin{customlegend}[legend columns = 1,legend style = {column sep=1ex}, legend cell align = left,
		legend entries={SS Fast, SS Quality, EdgeBoxes, CNN Threshold, FCN Threshold, TM Threshold, CNN Ranking, FCN Ranking, TM Ranking}]
		\addlegendimage{red,solid}
		\addlegendimage{red,dashed}		
		\addlegendimage{black,solid}
		\addlegendimage{blue,solid}
		\addlegendimage{blue,dashed}
		\addlegendimage{blue,densely dotted,thick}
		\addlegendimage{brown,solid}
		\addlegendimage{brown,dashed}
		\addlegendimage{brown,densely dotted,thick}
		\end{customlegend}
		\end{tikzpicture}
	\end{subfigure}
	\caption{Effect of the number of proposals on recall for different techniques. State of the art detection proposals methods can achieve high recall but only outputting a considerable number of proposals. Our proposed methods achieve high recall with orders of magnitude less output proposals.}
	\label{numProposalsVsRecall}
\end{figure*}
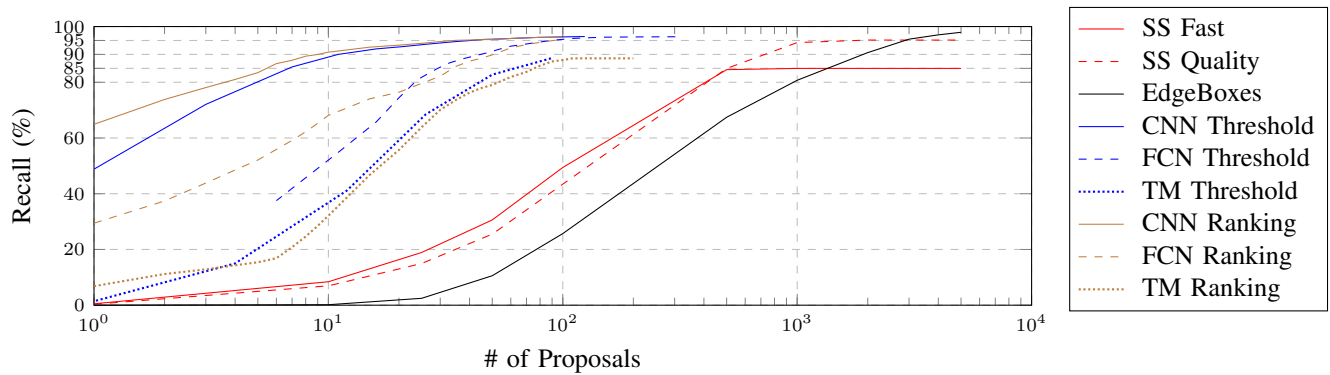

\textbf{Number of Proposals vs Recall}. Fig. \ref{numProposalsVsRecall} shows a recall comparison between all evaluated methods as we vary the number of output detection proposals. The best compromise between high recall and low number of proposals is CNN with objectness ranking. Cross-Correlation Template Matching performs poorly, requiring more proposals than our methods, but not reaching high recall, saturating at $88.59$ \%. All four methods proposed in this paper reach similar recall values (around $95$ \%) at 100 proposals per image. In comparison, EdgeBoxes and Selective Search requires one order of magnitude more proposals to produce similar recall.

It is also notable that CNN with objectness ranking can achieve $90$ \% recall with only 10 proposals per image, while FCN requires around 50 to reach similar recall. No other technique that we have evaluated can reach such high recall less than 50 proposals per image.

\begin{figure*}[ht]
	\centering	
	\begin{subfigure}[b]{0.36 \textwidth}
		\includegraphics[width=0.32\textwidth]{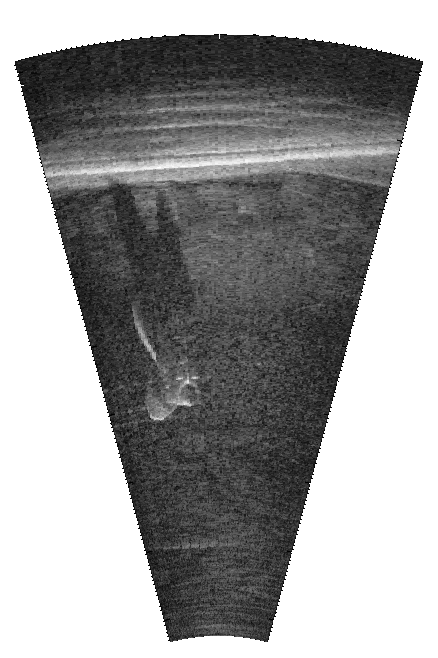}
		\includegraphics[width=0.32\textwidth]{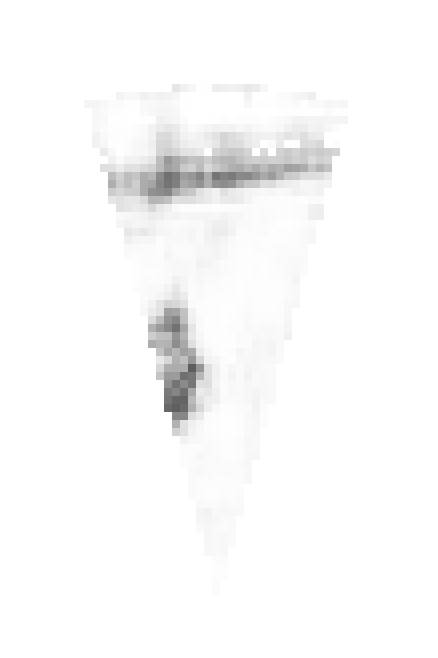}
		\includegraphics[width=0.32\textwidth]{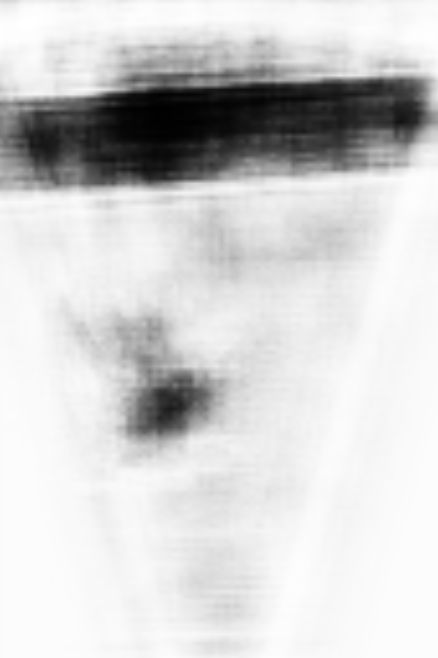}
		\caption{Wall and Propeller}
	\end{subfigure} \hspace*{1.5cm}
	\begin{subfigure}[b]{0.46 \textwidth}
		\includegraphics[width=0.32\textwidth]{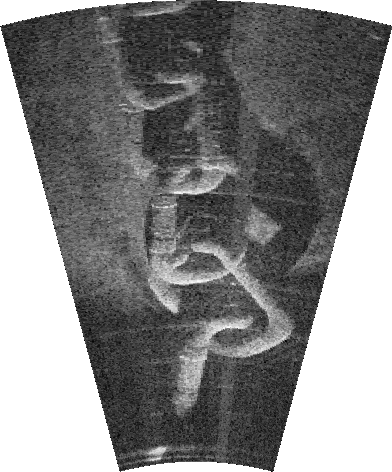}
		\includegraphics[width=0.32\textwidth]{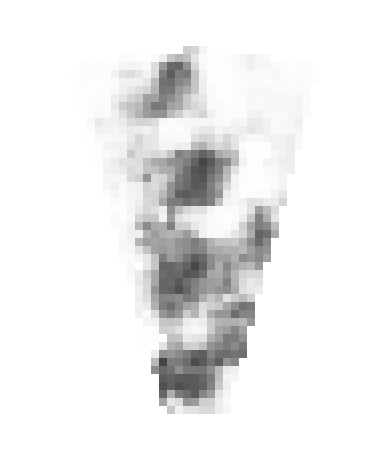}
		\includegraphics[width=0.32\textwidth]{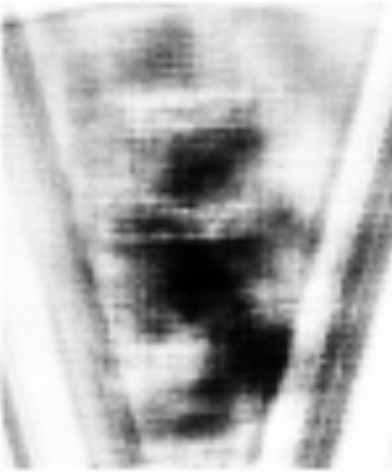}
		\caption{Chain}
	\end{subfigure}

	\begin{subfigure}[b]{0.40 \textwidth}
		\includegraphics[width=0.32\textwidth]{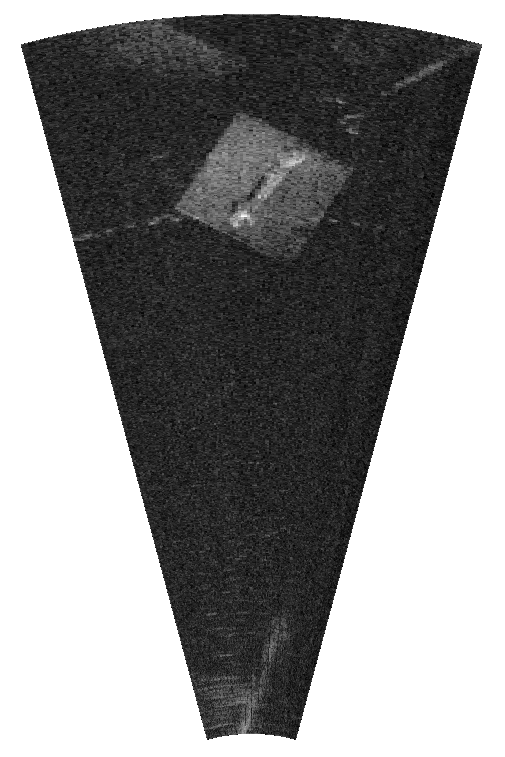}
		\includegraphics[width=0.32\textwidth]{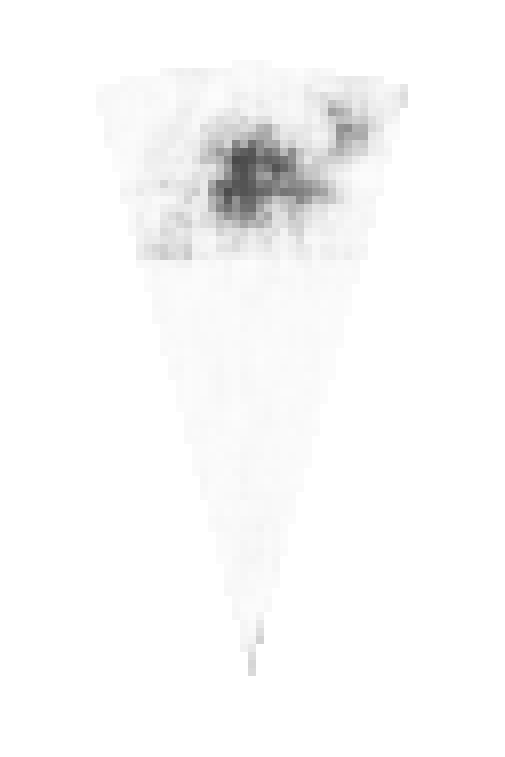}
		\includegraphics[width=0.32\textwidth]{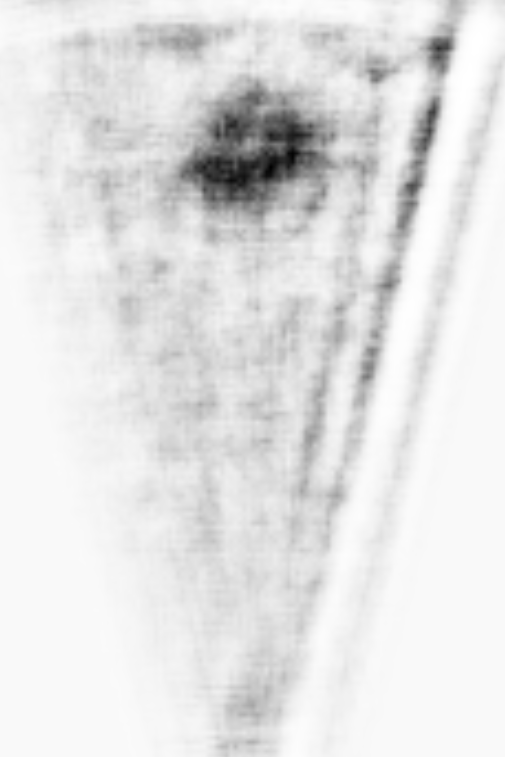}
		\caption{Rotating Platform with a Wrench}
	\end{subfigure} \hspace*{1.5cm}
	\begin{subfigure}[b]{0.40 \textwidth}
		\includegraphics[width=0.32\textwidth]{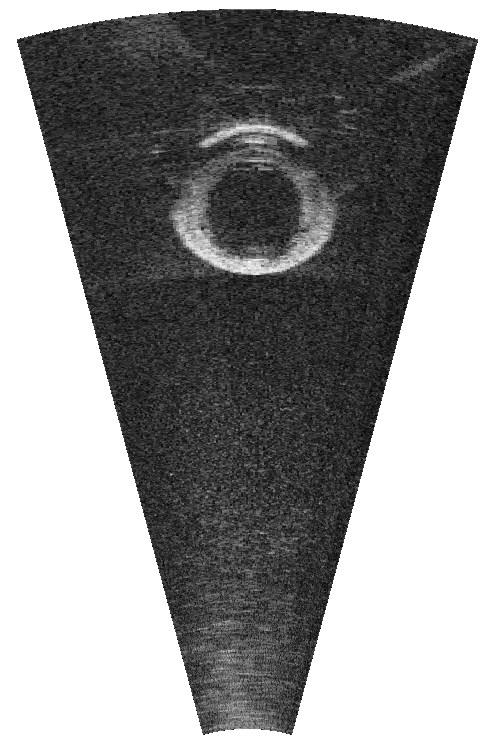}
		\includegraphics[width=0.32\textwidth]{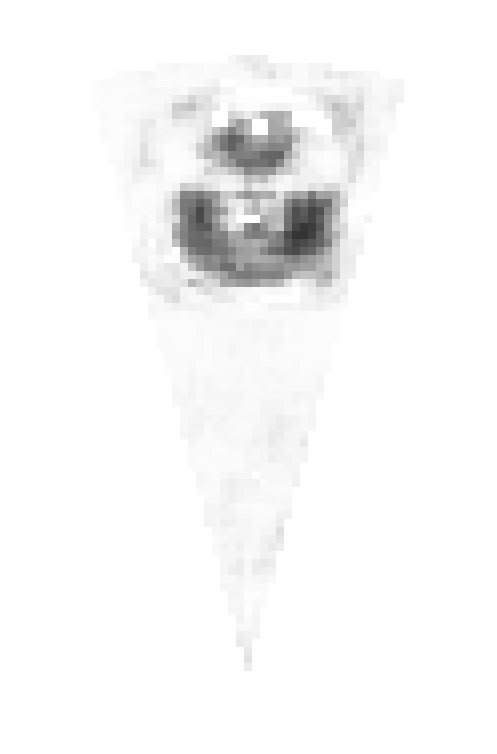}
		\includegraphics[width=0.32\textwidth]{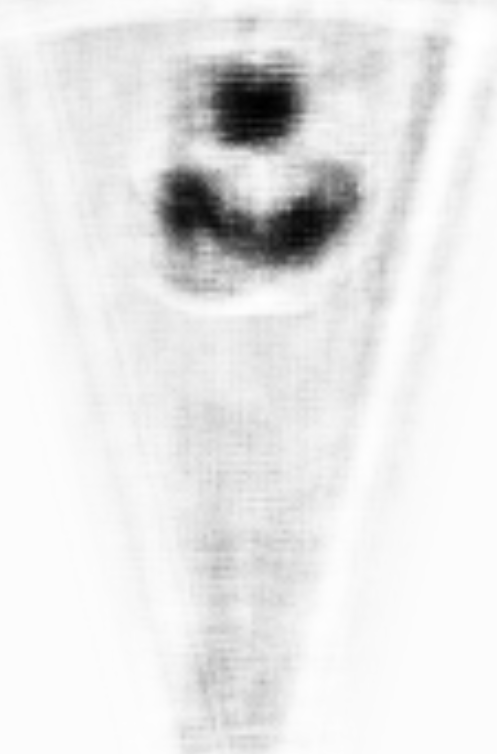}
		\caption{Large Tire}
	\end{subfigure}
	\caption{Visualization of objectness maps produced by CNN and FCN on previously unseen Forward-Looking Sonar Images. In each group: Left is the input image, Center is the CNN objectness map, while Right is the FCN map. Light shades represent low objectness, while Dark ones is high objectness.}
	\label{unseenVisualization}
\end{figure*}

\textbf{Generalization}. We now showcase the generalization ability of our proposed methods. For this we use three images that contain unseen objects, namely a Wall, Chain \footnote{This image was captured by CIRS, University of Girona.}, and a rotating platform with a Wrench. We visualize the objectness maps produced by CNN and FCN, which shows the spatial correlation between object position and objectness. Results are shown in Fig. \ref{unseenVisualization}. The wall in (a) shows a very good correlation with a high objectness, indicating that both CNN and FCN can produce detections over the wall, even as there is no wall example in the training set. Same effect can be seen in the Chain and Rotating Platform images. We also observe that CNN produces slightly lower objectness values than FCN, but both produce scores that can be easily distinguished from the background.

\section{Conclusions}

This work covers the problem of detecting novel objects without class information, which is applicable to the detection of hard to model objects such as marine debris underwater. We have shown a new fully convolutional network to estimate objectness maps from sonar images, and we have proposed objectness ranking to obtain detection proposals from objectness scores.

Our results on a marine debris dataset on Forward-Looking sonar images show that our methods can achieve high recall ($95$ \%) with a low number of output detections (80-300). In comparison EdgeBoxes \cite{zitnick2014edge} requires 5000 proposals to obtain $97$ \% recall, and Selective Search \cite{uijlings2013selective} needs around 2000 proposals to obtain $95$ \% recall. A baseline using classic cross-correlation template matching \cite{hurtos2013automatic} fails to generalize well and it considerably slower than the novel approaches we propose. These results show that a neural network can learn to predict appropriate objectness values efficiently, while generalizing to completely new objects.

We expect that our results will drive the development of new object detection techniques for sonar images, adding to new capabilities such as finding novel objects, or making underwater robots aware of unknown objects in the environment.

Future work includes evaluating other sensor modalities like side-scan or synthetic aperture sonar, and developing classifiers that can deal with information of unknown object classes.

\section*{Acknowledgements}

This work has been partially supported by the FP7-PEOPLE-2013-ITN project ROBOCADEMY (Ref 608096) funded by the European Commission, and by the Autonomous Harbour Cleaning project funded by EIT Digital (Ref 18181). The authors would like to thank Leonard McLean for his help in capturing data used in this paper.

\bibliographystyle{IEEEtran}
\bibliography{sonar-proposals-paper}

\begin{thebibliography}{10}
\providecommand{\url}[1]{#1}
\csname url@samestyle\endcsname
\providecommand{\newblock}{\relax}
\providecommand{\bibinfo}[2]{#2}
\providecommand{\BIBentrySTDinterwordspacing}{\spaceskip=0pt\relax}
\providecommand{\BIBentryALTinterwordstretchfactor}{4}
\providecommand{\BIBentryALTinterwordspacing}{\spaceskip=\fontdimen2\font plus
\BIBentryALTinterwordstretchfactor\fontdimen3\font minus
  \fontdimen4\font\relax}
\providecommand{\BIBforeignlanguage}[2]{{%
\expandafter\ifx\csname l@#1\endcsname\relax
\typeout{** WARNING: IEEEtran.bst: No hyphenation pattern has been}%
\typeout{** loaded for the language `#1'. Using the pattern for}%
\typeout{** the default language instead.}%
\else
\language=\csname l@#1\endcsname
\fi
#2}}
\providecommand{\BIBdecl}{\relax}
\BIBdecl

\bibitem{hurtos2013automatic}
N.~Hurt{\'o}s, N.~Palomeras, S.~Nagappa, and J.~Salvi, ``Automatic detection of
  underwater chain links using a forward-looking sonar,'' in
  \emph{OCEANS-Bergen, 2013 MTS/IEEE}.\hskip 1em plus 0.5em minus 0.4em\relax
  IEEE, 2013, pp. 1--7.

\bibitem{sawas2010cascade}
J.~Sawas, Y.~Petillot, and Y.~Pailhas, ``Cascade of boosted classifiers for
  rapid detection of underwater objects,'' in \emph{Proceedings of the European
  Conference on Underwater Acoustics}, 2010.

\bibitem{valdenegro2016submerged}
M.~Valdenegro-Toro, ``{S}ubmerged {M}arine {D}ebris {D}etection with
  {A}utonomous {U}nderwater {V}ehicles,'' in \emph{International Conference on
  Robotics and Automation for Humanitarian Applications (RAHA)}.\hskip 1em plus
  0.5em minus 0.4em\relax IEEE, 2016.

\bibitem{hosang2014good}
J.~Hosang, R.~Benenson, and B.~Schiele, ``How good are detection proposals,
  really?'' \emph{arXiv preprint arXiv:1406.6962}, 2014.

\bibitem{valdenegroToro2016Proposals}
M.~Valdenegro-Toro, \emph{Objectness Scoring and Detection Proposals in
  Forward-Looking Sonar Images with Convolutional Neural Networks}.\hskip 1em
  plus 0.5em minus 0.4em\relax Springer International Publishing, 2016.

\bibitem{valdenegro2016end}
------, ``{E}nd-to-{E}nd {O}bject {D}etection and {R}ecognition in
  {F}orward-{L}ooking {S}onar {I}mages with {C}onvolutional {N}eural
  {N}etworks,'' in \emph{Autonomous Underwater Vehicles (AUV), 2016
  IEEE/OES}.\hskip 1em plus 0.5em minus 0.4em\relax IEEE, 2016, pp. 144--150.

\bibitem{endres2010category}
I.~Endres and D.~Hoiem, ``Category independent object proposals,'' in
  \emph{Computer Vision--ECCV 2010}.\hskip 1em plus 0.5em minus 0.4em\relax
  Springer, 2010, pp. 575--588.

\bibitem{alexe2012measuring}
B.~Alexe, T.~Deselaers, and V.~Ferrari, ``Measuring the objectness of image
  windows,'' \emph{Pattern Analysis and Machine Intelligence, IEEE Transactions
  on}, vol.~34, no.~11, pp. 2189--2202, 2012.

\bibitem{zitnick2014edge}
C.~L. Zitnick and P.~Doll{\'a}r, ``Edge boxes: Locating object proposals from
  edges,'' in \emph{Computer Vision--ECCV 2014}.\hskip 1em plus 0.5em minus
  0.4em\relax Springer, 2014, pp. 391--405.

\bibitem{uijlings2013selective}
J.~R. Uijlings, K.~E. van~de Sande, T.~Gevers, and A.~W. Smeulders, ``Selective
  search for object recognition,'' \emph{International journal of computer
  vision}, vol. 104, no.~2, pp. 154--171, 2013.

\bibitem{ren2015faster}
S.~Ren, K.~He, R.~Girshick, and J.~Sun, ``Faster r-cnn: Towards real-time
  object detection with region proposal networks,'' in \emph{Advances in Neural
  Information Processing Systems}, 2015, pp. 91--99.

\bibitem{girshick2015fast}
R.~Girshick, ``Fast r-cnn,'' in \emph{Proceedings of the IEEE International
  Conference on Computer Vision}, 2015, pp. 1440--1448.

\bibitem{lecun1998gradient}
Y.~LeCun, L.~Bottou, Y.~Bengio, and P.~Haffner, ``Gradient-based learning
  applied to document recognition,'' \emph{Proceedings of the IEEE}, vol.~86,
  no.~11, pp. 2278--2324, 1998.

\bibitem{iandola2016squeezenet}
F.~N. Iandola, S.~Han, M.~W. Moskewicz, K.~Ashraf, W.~J. Dally, and K.~Keutzer,
  ``Squeezenet: Alexnet-level accuracy with 50x fewer parameters and< 0.5 mb
  model size,'' \emph{arXiv preprint arXiv:1602.07360}, 2016.

\bibitem{ValdenegroToro2017RealtimeCN}
M.~Valdenegro-Toro, ``Real-time convolutional networks for sonar image
  classification in low-power embedded systems,'' \emph{CoRR}, vol.
  abs/1709.02153, 2017.

\bibitem{Shelhamer2015FullyCN}
E.~Shelhamer, J.~Long, and T.~Darrell, ``Fully convolutional networks for
  semantic segmentation,'' \emph{2015 IEEE Conference on Computer Vision and
  Pattern Recognition (CVPR)}, pp. 3431--3440, 2015.

\bibitem{kingma2014adam}
D.~Kingma and J.~Ba, ``Adam: A method for stochastic optimization,''
  \emph{arXiv preprint arXiv:1412.6980}, 2014.

\bibitem{hosang2015makes}
J.~Hosang, R.~Benenson, P.~Doll{\'a}r, and B.~Schiele, ``What makes for
  effective detection proposals?'' 2015.

\bibitem{chavali2016object}
N.~Chavali, H.~Agrawal, A.~Mahendru, and D.~Batra, ``Object-proposal evaluation
  protocol is' gameable','' in \emph{Proceedings of the IEEE Conference on
  Computer Vision and Pattern Recognition}, 2016, pp. 835--844.

\bibitem{midelfart2009template}
H.~Midelfart, J.~Groen, and O.~Midtgaard, ``Template matching methods for
  object classification in synthetic aperture sonar images,'' in
  \emph{Proceedings of the Underwater Acoustic Measurements Conference}, no. S
  S, 2009.

\end{thebibliography}

\end{document}